# Medical Dataset Classification for Kurdish Short Text over Social Media

Ari M. Saeed [a,*], Shnya R. Hussein [b], Chro M. Ali [c], Tarik A. Rashid [d]

[a] Computer Science Department, University of Halabja, Halabja, Kurdistan, KRG, Iraq
[b] Computer Science Department, University of Halabja, Halabja, Kurdistan, KRG, Iraq
[c] Computer Science Department, University of Halabja, Halabja, Kurdistan, KRG, Iraq
[d] Computer Science and Engineering Department, University of Kurdistan-Hawlêr, Erbil, Kurdistan, KRG, Iraq

ari.said@uoh.edu.iq

## Abstract

The Facebook application is used as a resource for collecting the comments of this dataset, The dataset consists of 6756 comments to create a Medical Kurdish Dataset (MKD). The samples are comments of users, which are gathered from different posts of pages (Medical, News, Economy, Education, and Sport). Six steps as a preprocessing technique are performed on the raw dataset to clean and remove noise in the comments by replacing characters. The comments (short text) are labeled for positive class (medical comment) and negative class (non-medical comment) as text classification. The percentage ratio of the negative class is 55% while the positive class is 45%.

## Specifications Table

| | |
|---|---|
| **Subject** | Applied Machine Learning |
| **Specific subject area** | Medical Dataset Classification for Kurdish Short Text over Social Media |
| **Type of data** | Text<br>Figure<br>Table |
| **How the data were acquired** | Facepager application is used for collecting the comments after configuring. |
| **Data format** | Raw |
| **Description of data collection** | Each post is separated accurately to describe the type of class (medical or non-medical), then the link of the post is copied and pasted in the Facepager application for gathering the specified comments. |
| **Data source location** | Kurdish post link in Facebook Application |
| **Data accessibility** | +Repository name: Mendeley Data<br>Data identification number: 10.17632/f2yfz4r9fr.1<br>Direct link to the dataset: https://dx.doi.org/10.17632/f2yfz4r9fr.1 |

## Value of the Data

1- This is an effort of collecting a dataset in the field of medical text classification for the Kurdish language. Moreover, It can be beneficial for supporting and modeling patient health systems, health policies, and regulations.
2- The data is preprocessed and ready for implementation by those researchers and scholars who conduct research work on the Arabic Alphabet, such as Persian, Arabic, and Urdu.
3- The dataset can be used with several preprocessing steps such as stemming and lemmatization.

## Data Description

The Kurdish language is one of the languages of the Middle East that is used for speaking by Kurdish people. Central Kurdish (Sorani) and Kurmanji are two popular dialects of the Kurdish Language [1, 2]. In this project, the Sorani dialect is used for collecting the database. The texts of Sorani Kurdish are like Persian, Arabic, and Urdu that are written from right to left. The characters among those languages are almost like each other but sometimes have different Unicode as shown in Table 1. The number of Sorani Kurdish alphabets is 36 that is divided into vowels (ا, ە, و, ۆ, وو, ی, ێ) and consonants (ئ, ب, پ, ت, ج, چ, ح, خ, د, ر, ڕ, ز, ژ, س, ش, ع, غ, ف, ڤ, ق, ک گ, ل, ڵ, م, ن, ه, و, ی) [3, 4]. The character (و, ی) are used as vowels and constants based on the positions of the word, for example, the word (گوڵ) (gull) means (Flower), the (و) is a vowel, while the word (وازی) (wazi) means (game), the (و) is constant, the word (یاری) (yari) means (play), and the first (ی) is constant, by contrast, the second one is a vowel. The Kurdish language is complex and has different scripts (no standard) for Sorani, dialect for example, in some sources (ك) is used instead of (ک) [3, 4].

*Table 1 Alphabet similarities among (Kurdish, Arabic, Persian, Urdu) Languages*

| NO. | Kurdish alphabetic | Arabic Language | Persian Language | Urdu Language |
|---|---|---|---|---|
| 1 | ئ | ا | آ orا | آ orا |
| 2 | ب | ب | ب | ب |
| 3 | پ | ت | پ | پ |
| 4 | ت | ث | ت | ت |
| 5 | ج | ج | ث | ٹ |
| 6 | چ | ح | ج | ث |
| 7 | ح | خ | چ | ج |
| 8 | خ | د | ح | چ |
| 9 | د | ذ | خ | ح |
| 10 | ر | ر | د | خ |
| 11 | ڕ | ز | ذ | د |
| 12 | ز | س | ر | ڈ |
| 13 | ژ | ش | ز | ذ |
| 14 | س | ص | ژ | ر |
| 15 | ش | ض | س | ڑ |
| 16 | ع | ط | ش | ز |
| 17 | غ | ظ | ص | ژ |
| 18 | ف | ع | ض | س |
| 19 | ڤ | غ | ط | ش |
| 20 | ق | ف | ظ | ص |
| 21 | ک | ق | ع | ض |
| 22 | گ | ك | غ | ط |
| 23 | ل | ل | ف | ظ |
| 24 | ڵ | م | ق | ع |
| 25 | م | ن | ک | غ |
| 26 | ن | ه | گ | ف |
| 27 | ه | و | ل | ق |
| 28 | و | ي | م | ک |
| 29 | ی | ء | ن | گ |
| 30 | ا | | و | ل |
| 31 | ە | | ه | م |
| 32 | و | | ی | ن |
| 33 | ۆ | | | و |
| 34 | وو | | | ه |
| 35 | ی | | | ی |
| 36 | ێ | | | ے |

## Data collection

In this era, the health of people is a serious subject that researchers work on it closely [7, 8]. For this purpose, it is important to read humans' views over social media. In this work, the Facebook application is used as a social media for creating a proper MKD. Nevertheless, to say that for predicting the right sight of humans by using machines, a good resource (dataset) is necessary. As it is clear, there are so many channels, websites, and live posts that can be used for this purpose. The database in this work id consisted of 6756 samples, which are divided into two different classes (medical and non-medical). The samples were collected from various pages and different areas as shown in Table 2. The number of medical comments (positive class) is 3076 while the non-medical comments (negative class) are 3680.

*Table 2 Number and Percentage of Collected Dataset*

| Class | Field | No. of Samples | Percentage |
|---|---|---|---|
| Medical | Medical | 3076 | 45% |
| Not Medical | News | 890 | 55% |
| | Economy | 720 | |
| | Education | 1140 | |
| | Sport | 930 | |

## Methodology

On social media, the data can be viewed in various types, such as image, video, text. In this work, the data set is collected from the text. Facebook application is used for collecting the comments of users. Some different tools and techniques can be utilized for collecting the comments, the Facepager tool is one of them that has been used for this reason [9]. The following steps should be followed for obtaining the data as shown below in Figure 1.

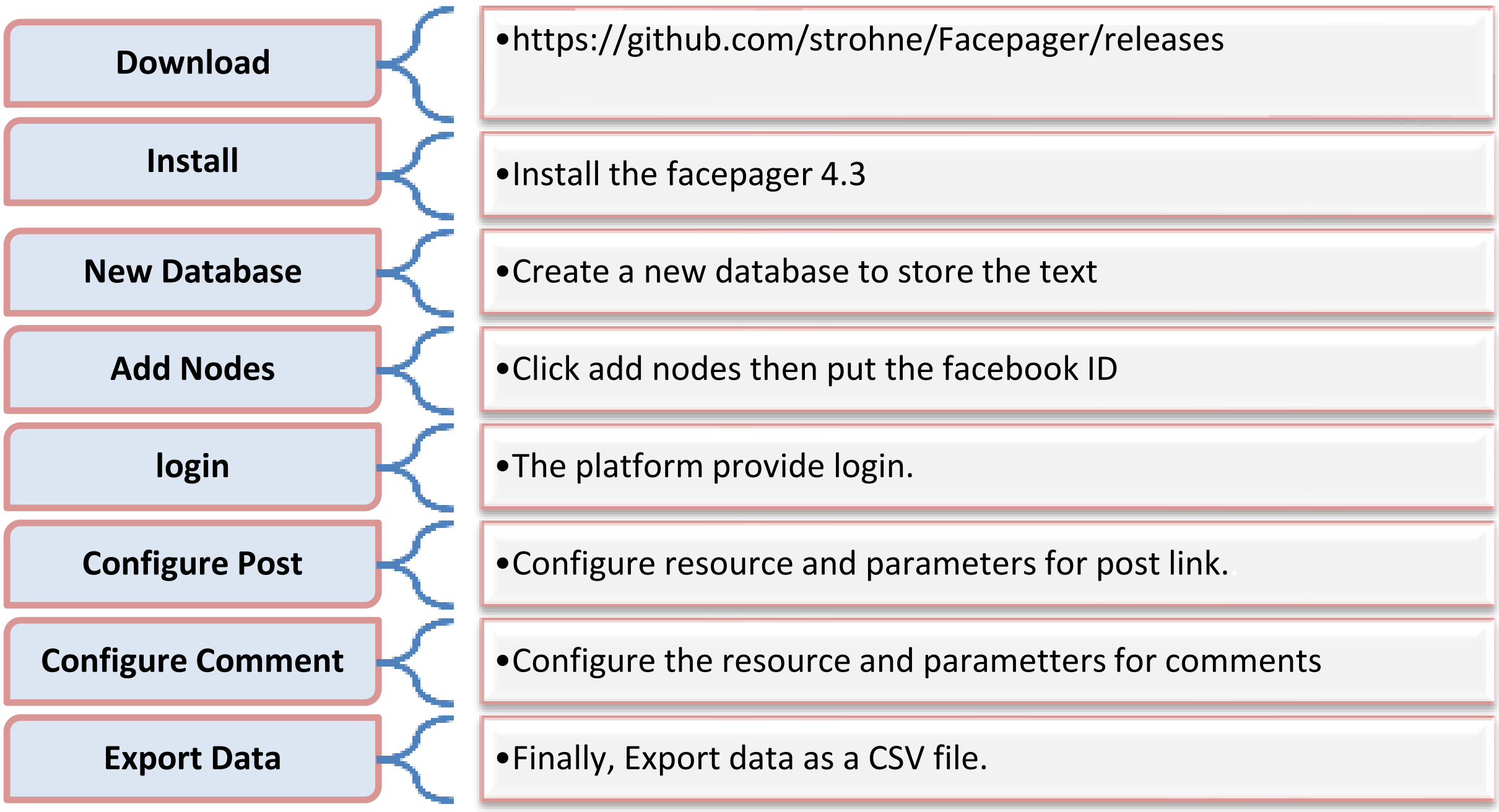

*Figure 1 Steps to Dataset Collection*

As shown in Figure 1, the first step is downloading the Facepager software for collecting the comments. The second step is locating and installing the files. The third step is to open the software and create a new database for saving the text file in (.db) format. The fourth step is adding nodes and putting the Facebook ID of the specified link after converting it over the internet.  The fifth step is to log into Facebook via the Facepager tool. The sixth important step is configuring resources as (/<page-id>/posts) and parameters filed as (message) and specifying a start date and end date to fetch posts between those specific dates as shown in Figure 2.

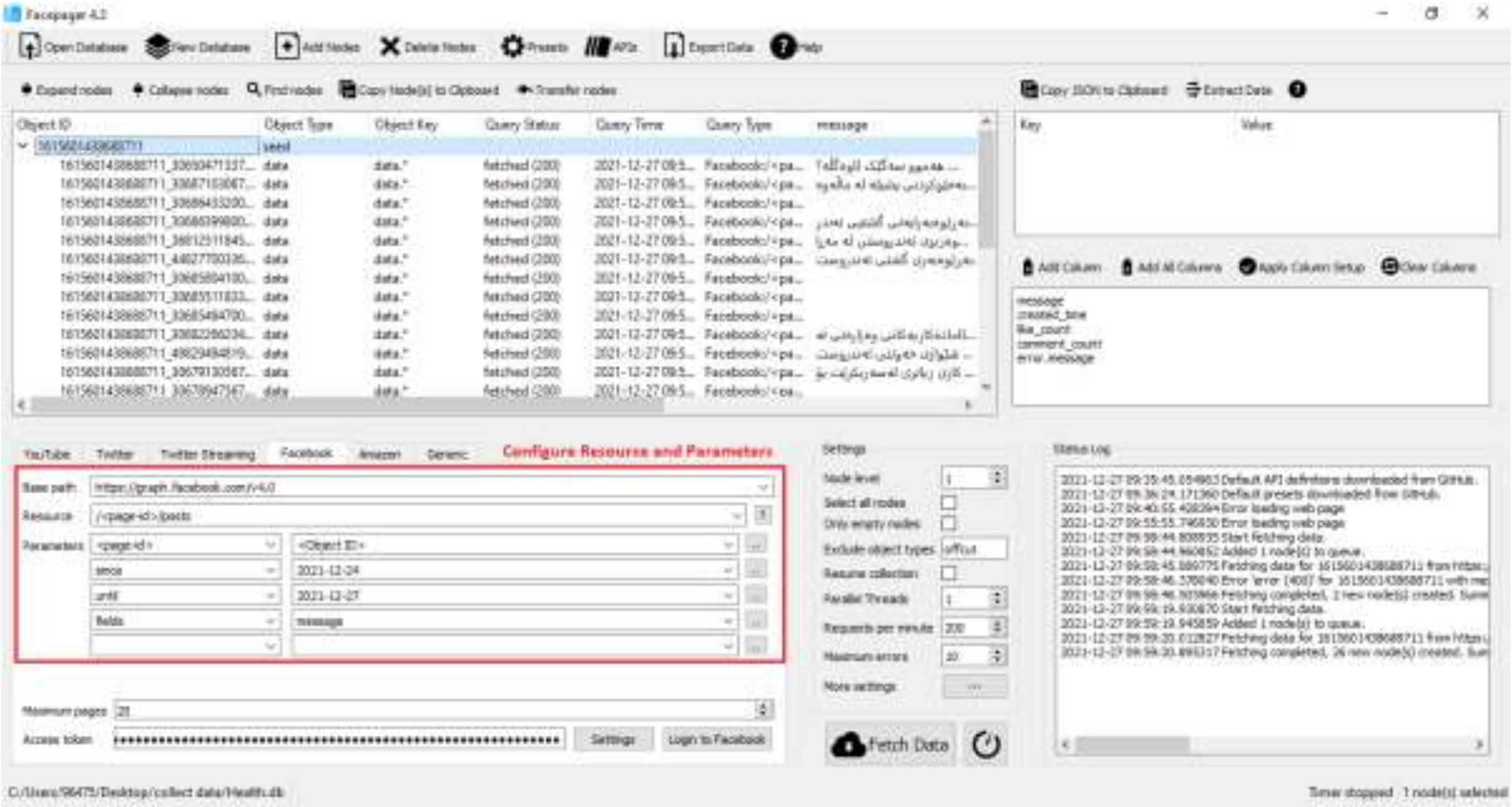

*Figure 2 Configure Facepager Posts*

The seventh step is configuring a tool for fetching comments by clicking on a specific post and configuring resources as (/<post-id>/comments) and parameters filed as (message) as shown in Figure 3.

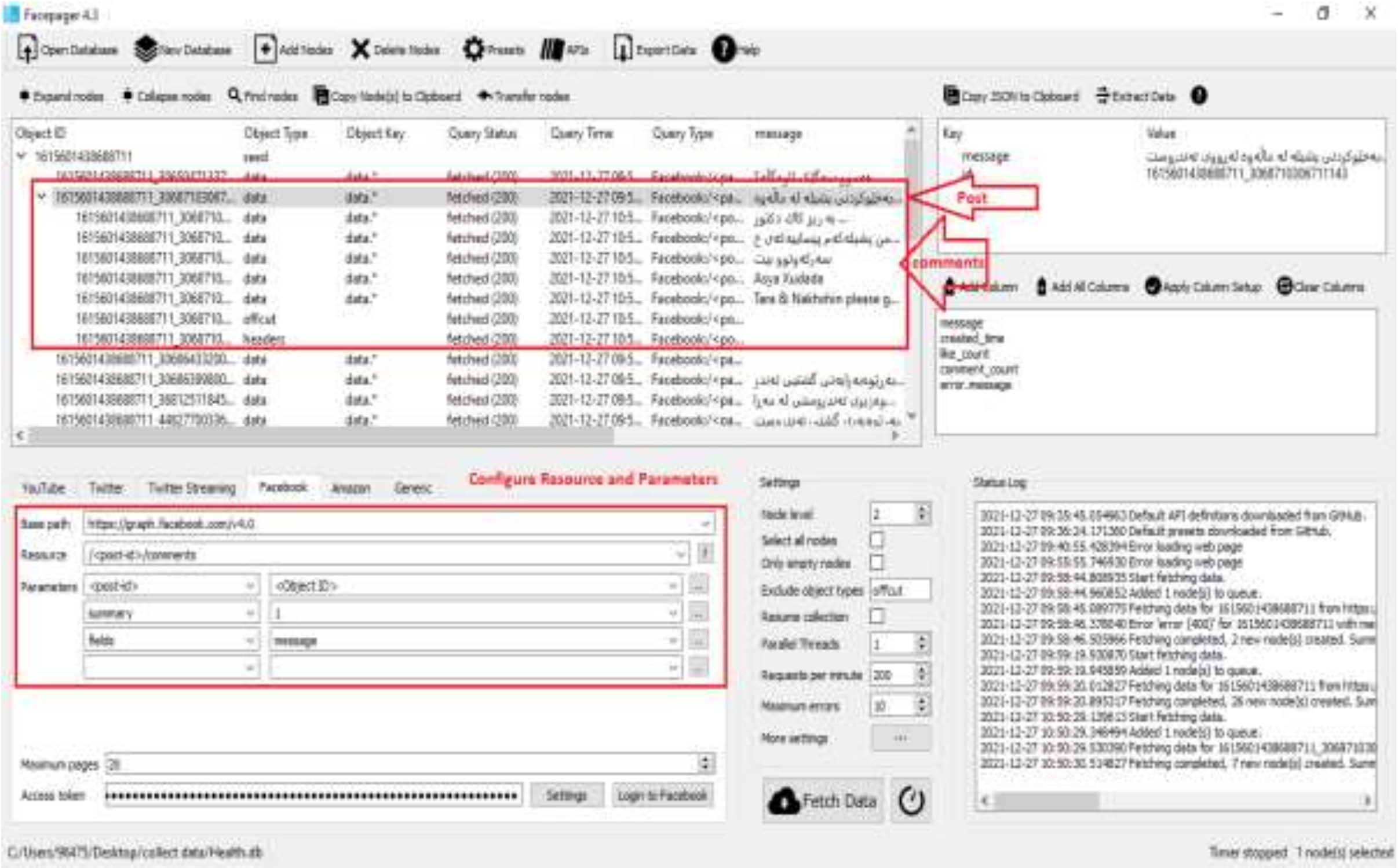

*Figure 3 Configure Facepager Comments*

The last and final step is exporting the comments as a CSV file as shown in Figure 4.

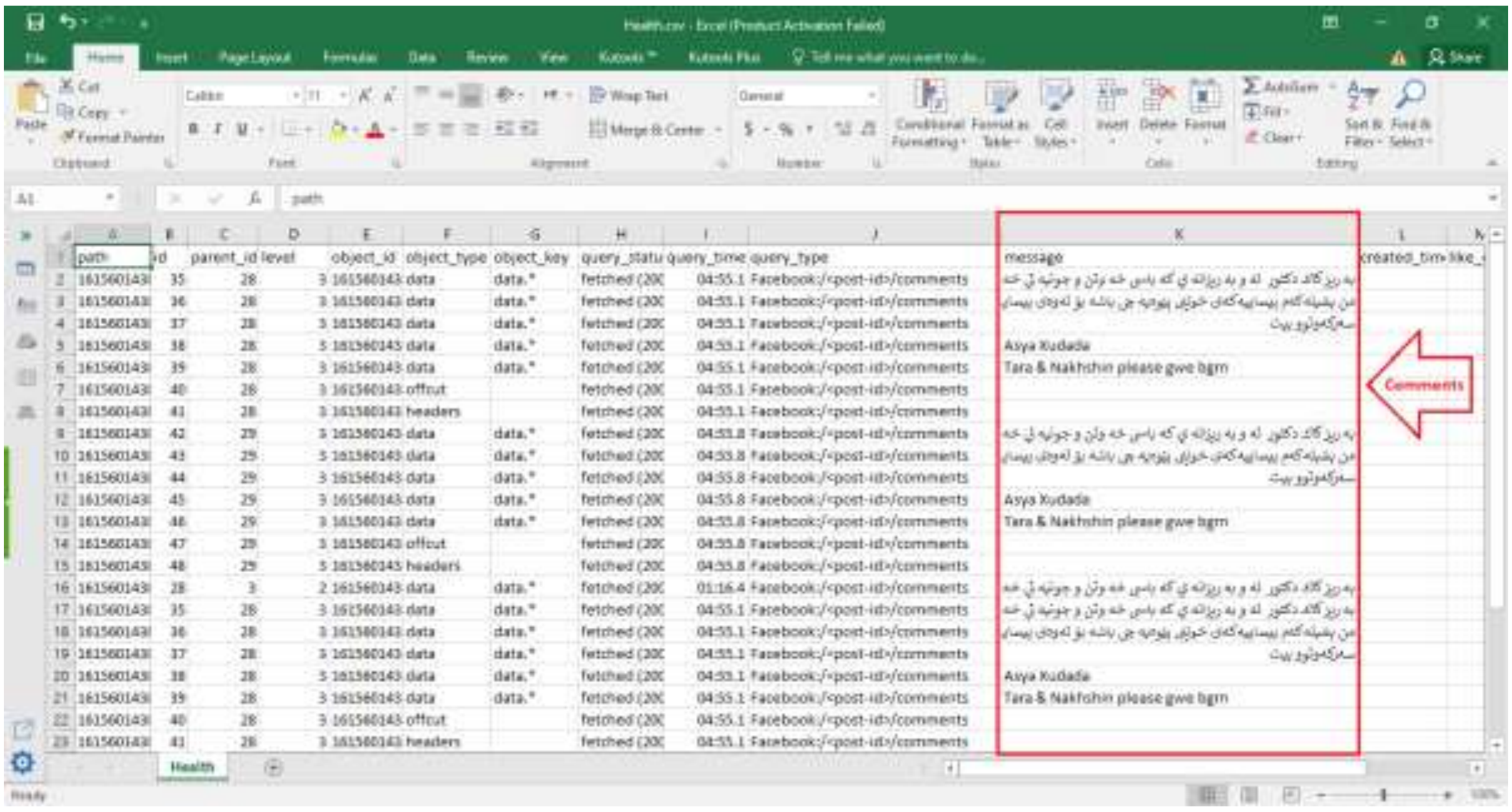

Figure *4 Configure Facepager Comments*

# Data set preprocessing

Preprocessing is one of the most important challenges for decreasing the noise on social media. Due to Kurdish users on the Facebook application using different Unicode to share their opinion and views. This causes a big issue for recognizing text and makes different characters shape. Using different scripts also increases the number of features (word) [1, 4, 5, and 10]. Accordingly, python language is used to create a new tool for implementing the below steps on the text as shown in Table 3:

1. Removing noise (URL, User mentions, and Hashtag) on social media users will provide extra information for their relatives and friends by using URL, mentions (@user name), and hashtags (#special topic) that information are helpful for users but it is noise for the machine. It has to be removed.
2. Replacing elongated characters: users on social media sometimes use elongated words purposely to emphases about special things, such as (جييييييـــة) (chiye), which means (Whaaaaaaat), which should be replaced with a base word ( جية) (chiye), which means (what).
3. Incorrect spelling and grammar: sometimes it is easy for users to correct the misspelling and grammar but machines cannot understand and it is challenging. These three words (ماشـــاءاللّا, ماشــــاء الله ) (masha allh), which means (Allah has willed it) used as a misspelling instead the correct word (ماشاالله), which means (Allah has willed it).
4. Removing punctuations: users on social media use them to express special emotions, which are easy for a human to recognize. Nevertheless, those punctuations make usefulness for machines to translate and become inefficient for text classification. These punctuations are removed (!@#$%^&*()[]{};:,٫٠./<>؛ ؟٠!?× ٠٪!! ،‘،،،؛ £,’~ِـــــــــــــــ،؛‘ٍَُّ ٓ\ ""''+_=-~`|?؟ ”,*ٰٖٗ٘ٙٚٛٝٞ ۖ ۗ ۘ ۙ ۚ ۛ ۜ ٜ ؛ ۞ ۭ…©).

5. Removing numbers: numbers increase the number of features in text datasets on social media and they are not helpful for the machine to understand. However. Kurdish users use different types of numbers, such as (English, Arabic, and Kurdish) numbers as shown.
6. Replacing characters: due to the Kurdish language using the same script of Arabic language for some characters and some users on social media use Arabic Keyboard for writing. This has become an issue for matching and selecting features. However, the issue has been solved by replacing the character as shown below:
   a- ‘ي’ with ‘ی’
   b- ‘ك’ with ‘ک’
   c- ‘ۇ’ with ‘ۆ’
   d- ‘ة’ with ‘ه’
   e- If the word ends with ‘ئ’ replace with ‘ئ’
   f- If the word ends with ‘ه’ (\u 647\u 200C), then replace it with ‘ه’ (\u0647) as shown in the same shape of characters but different Unicode.

*Table 3 Preprocessing Steps*

| NO | Preprocessing Steps | Natural Comments with the Arabic alphabet | Natural Comments with Latin alphabet | Natural Comments in English | Preprocessing Comments with the Arabic alphabet | Preprocessing Comments with Latin alphabet | Preprocessing Comments in English |
|---|---|---|---|---|---|---|---|
| 1. | Removing noise (URL, User mentions and Hashtag): | دکتۆره سەناریا زیاد زەنگەنە. پسپۆری نەشتەرگەری و نەخۆشیەکانی چاو. Medical_# Knowledge #زانستی_پزی شکی | dktore se nariya ziyad ze ngene. psporî ne şitergerî û nexoşiye kanî çaw. #Medical_ Knowledge #zanistî_p zîşkî | Doctor Sanaria Zyad Zangan. Expert in eye surgery and disease. Medical_K nowledge #زانستی_پزی شکی | دکتۆره سەناریا زیاد زەنگەنە پسپۆری نەشتەرگەری و نەخۆشیەکانی چاو | dktore se nariya ziyad zengene psporî neşite rgerî û ne xoşiyekanî çaw | Doctor Sanaria Zyad Zangan expert in eye surgery and disease |
| 2. | Replacing elongated characters | سلاو دکتور خوشکیکم هةية لة حاديسا سةري بيكراوة ايستا هةردوو بيلبيلةي جاوي بة باشي ناجوليني تكاييييييي | slaw dktur xuşkîkm heye le ḧadîsa serî bîkrawe aysta herdû bîlbîleyi cawî be | Hi doctor, my sister was hurt in an accident now her cornea cannot move perfectly | سلاو دکتور خوشکیکم هەیە لە حادیسا سەری بیکراوە ایستا هەردوو بیلبیلەی جاوی بە باشی ناجولینی تکایە جواب هوکار جیە | slaw dktur xuşkîkm heye le ḧadîsa serî bîkrawe aysta herdû bîlbîleyi cawî be başî naculînî tkaye cwab hukar ciye | Hi doctor, my sister has hurt in an accident now her cornea cannot move perfectly please answer and what is the cause |
| | | بیبیة جوابببببب هوکار جیییییبیة | başî naculînî tkayîyîyîyîy îyîyîyîye cwabibbib bibbibb hukar cîyîyîyîye | pleaaaaaa aaaaase answeeee eer and whaaaaaa at is the cause | | | |
| 3. | Incorrect spelling and grammars | ماشاڵلا هەرچەند جار ئەو کابرایە باسی بەنزین بکات گرانتر دەبێت | maşallla herçend car ew kabraye basî benzîn bikat grantir debêt | Allah has willed it, any time that man talks about gasoline, it will be more expensive | ماشاالله هەرچەند جار ئەو کابرایە باسی بەنزین بکات گرانتر دەبێت | maşaalle herçend car ew kabraye basî benzîn bikat grantir debêt | Allah has willed it, how any time that man talks about gasoline, it will be more expensive |
| 4. | Removing punctuation | بەقسەی تۆبێت ئەم وڵاتە شامی شەریفە!!!! | beqseyi tobêt em wullate şamî şerîfe!!!! | According to your speech, this country is peaceful!!!! | بەقسەی تۆبێت ئەم وڵاتە شامی شەریفە | beqseyi tobêt em wullate şamî şerîfe | According to your speech, this country is peaceful |
| 5. | Removing numbers: | دکتورە من ٣٠ ساله موي ردینم دە ر دینم هە ر دیتە وە | dkturh min 30 salh mwî rdînm dh r dînm hh r dîth wh | Doctor, it is about 30 years, I have been pulling out beard hair, yet it grows back | دکتورە من ساله موی ردینم دە ر دینم هە ر دیتە وە | dkturh min salh mwî rdînm dh r dînm hh r dîth wh | Doctor, it is about 30 years, I have been pulling out beard hair, yet it grows back |
| 6. | Replacing characters | سلاو من حساسیةتی زۆری چاوم هةیة بةتایبةتی لةوةرزی بةهار سوردةبیتةوةو دةخوریت ئایا چارةسةری بنةرةتی هةیة یان رینمایی بیزةحمةت؟ | slaw min ḣsasiyetî zğrî çawm heye betaybetî lewerizî behar surdebîte wew dexurît aya careserî bineretî heye yan rînmayî bîzeḣmet? | Hi, I have an eye rash, especially in the Spring season and it became red and itchy, is there any essential treatment or any advice. | سلاو من حساسیەتی زۆری چاوم هەیە بەتایبەتی لەوەرزی بەهار سوردەبیتەوەو دەخوریت ئایا چارەسەری بنەرەتی هەیە یان رینمایی بیزەحمەت | slaw min ḣsasiyetî zorî çawm heye betaybetî lewerizî behar surdebîtewe w dexurît aya çareserî bineretî heye yan rînmayî bîzeḣmet | Hi, I have an eye rash, especially in the Spring season and it became red and itchy, is there any essential treatment or any advice. |

## Dataset Labeling:

After collecting the dataset, another important step is labeling the samples. For this purpose, three annotators read the samples accurately and manually labeled the unlabeled samples for two classes (medical and non-medical). This process needs a huge effort and consumes time. For labeling each sample, the annotator annotates the sample based on some special words in the medical domain and the meaning of each sentence as shown in Table 4:

*Table 4 Labeling of Comments*

| NO. | Samples (Comments) in the Arabic alphabet | Samples (Comments) in Latin alphabet | Samples (Comments) in English | Classes |
|---|---|---|---|---|
| 1 | سلاو من ده مو جاو م خالى قاوەى يه جى باشه بيزه حمه ت | slaw min de mu caw m xalî qaweyi ye cî başe bîze ḣme t | Hello, my face has a brown spot, please what is good for me to do | medical |
| 2 | پۆلێک ژمارەى ته له به كانى ٦٠ بۆ سه رەوه بێ ئه وانه هه مووى بێ مانان وەهيچ سوودى نيه! | polêk jmareyi te lh bih kanî ٦٠ bo sh rewe bê e wanh he muwî bê manan wehîç sûdî niye! | A class that has several students, more than 60, that is no sense and does’t have any benefit | Not medical |
| 3 | سلاو دكتور مناله كه م ماوه يه سكى ئه جيت لينجى بيوه شير ى قتو ئه خوات | slaw dktur minalh kh m mawh yh skî ih cît lîncî biyuh şîr yi qtu ih xwat | Hi doctor, my baby has diarrhea and viscidity and eats condensed milk | medical |
| 4 | ئەو رۆژەی بەخت دەرگایت لێ ئەکاتەوە بەلام ئەقل دایئەخاتەوە | ew rojeyi bext dergayt lê ekatewe belam eql dayexatewe | The day that opens the luck for you, yet, it closes mind | Not medical |

# Ethics Statement

All omments in the dataset belong to users in the Facebook application and it is scrapped. The data has been distributed over Facebook and thus, it has been collected and labeled. Moreover, we confirm that all the data is insensitive and anonymized data

# .Declaration of Competing Interest

The authors declare that they have no known competing financial interests or personal relationships which have, or could be perceived to have, influenced the work reported in this article.

# CRediT Author Statement

Ari M. Saeed: Supervision, Data curation, Conceptualization, Methodology, Visualization, Project administration, Funding acquisition, Writing - Original Draft, Writing –review and editing; Shnya R. Hussein: Software, Formal analysis, Investigation, and Resources; Chro M. Ali: Software, Formal analysis, Investigation, and Resources; Tarik A. Rashid: Methodology, Supervision, Validation, Writing –review & editing.

# Acknowledgments

The authors would like to thank the University of Halabja and the University of Kurdistan Hewler for providing all the facilities needed for conducting this research work.